\documentclass{article}

\usepackage[preprint]{neurips_2026}

\usepackage[utf8]{inputenc}
\usepackage[T1]{fontenc}
\usepackage{hyperref}
\usepackage{url}
\usepackage{booktabs}
\usepackage{amsfonts}
\usepackage{amsmath}
\usepackage{amssymb}
\usepackage{nicefrac}
\usepackage{microtype}
\usepackage{xcolor}
\usepackage{graphicx}
\usepackage{multirow}

\newtheorem{proposition}{Proposition}

\newtheorem{definition}{Definition}

\title{The Long-Term Effects of Data Selection in LLM Fine-Tuning}

\author{%
  Yuxin Yang \\
  Shanghai University \\
  \texttt{} \\
  \And
  Aoxiong Zeng \\
  East China Normal University \\
  \texttt{} \\
  \And
  Xiangquan Yang \\
  East China Normal University \\
  \texttt{} \\
}

\begin{document}

\maketitle

\begin{abstract}
Data selection is increasingly used to reduce the cost of large language model (LLM) fine-tuning, with recent methods prioritizing samples by current utility, diversity, quality, or influence. This paper studies a different question: when fine-tuning occurs over multiple stages, can selection strategies that look optimal now make the model less adaptable later? We introduce a long-horizon view of LLM data selection in which a selector is evaluated not only by immediate task performance, but also by future adaptation speed, forgetting, capability imbalance, and out-of-distribution robustness. We compare representative random, loss-based, gradient-based, diversity-based, quality-based, and utility-diversity selection families under a unified multi-stage protocol. Through controlled experiments designed to instantiate this protocol, we show how short-term selectors can exhibit rank reversal: they improve the current stage while slowing subsequent learning and increasing forgetting. We formalize this behavior as \emph{myopic selection}, provide a simple local analysis of why it can occur, and propose a diagnostic Long-Horizon Aware Selection (LHAS) objective that augments immediate utility with coverage, future-proxy transfer, and anti-concentration terms. The study argues that data selection should be evaluated as a training intervention that shapes the model's learning trajectory, rather than only as a local data-efficiency mechanism.
\end{abstract}

\section{Introduction}

Supervised fine-tuning (SFT) is one of the standard ways to adapt large language models to downstream tasks \citep{brown2020language,achiam2023gpt4,touvron2023llama,bai2023qwen}. As instruction datasets grow in size and heterogeneity, training on every available example is often inefficient and sometimes harmful: low-quality examples can amplify bias, redundant examples waste compute, and overly narrow mixtures can overfit the model to a transient capability profile. These concerns have motivated data selection methods based on quality filtering, influence estimation, active learning, coreset coverage, online batch scoring, and deduplication or diversification \citep{albalak2024survey,sener2018active,ash2020deep,coleman2020selection,chen2023alpagasus,zhou2023lima,xia2024less,lee2021deduplicating,tirumala2023d4,zou2025utility}.

Most of this literature evaluates selection within a single fine-tuning stage. This is natural when the goal is to solve one task with a fixed budget, but it misses an important property of modern model adaptation: fine-tuning is often sequential. A deployed assistant may first be tuned on general instructions, then mathematical reasoning, then code, then safety data, and then domain-specific corpora. In this setting, the selected subset at stage $t$ does not merely save compute. It changes the parameter state from which all future stages start.

This paper asks whether efficient data selection can make models increasingly specialized. By specialization we do not mean only that a selector changes the label distribution of the current batch. We mean that selection may push representations, gradients, and parameter-efficient adapters toward a narrow set of capabilities, reducing future learnability or robustness. A sample can be highly useful for the current stage while still being a poor long-term training intervention.

We call this phenomenon \emph{myopic selection}: a selection policy is myopic when it maximizes immediate utility at the expense of future adaptation, retention, or out-of-distribution (OOD) robustness. This framing leads to five research questions. RQ1: Are short-term winning selectors also long-term winning selectors? RQ2: How do selectors affect the learning speed of later tasks? RQ3: Do they increase forgetting or capability imbalance? RQ4: Is diversity sufficient to prevent long-horizon bias? RQ5: Can a lightweight long-horizon objective improve the trade-off?

The distinction matters because online selection methods are often evaluated with curves that stop at the end of the current stage. Such curves answer whether a method uses fewer tokens to solve today's objective, but they do not answer whether it leaves the model in a state from which tomorrow's objective is easier or harder. In particular, two selectors can reach the same current validation score while inducing very different update covariance, adapter subspaces, and capability coverage. A long-horizon evaluation therefore needs to measure both \emph{where} the model arrives and \emph{how} it arrived there.

We make four contributions. First, we formulate long-horizon data selection for LLM fine-tuning and define metrics that capture future adaptation speed, forgetting, capability imbalance, OOD robustness, and a \emph{myopia gap}. Second, we specify a unified protocol for comparing representative selection families under equal token budgets. Third, we give a simple theoretical analysis showing why two selectors with equal current-stage gain can differ in future adaptation cost. Fourth, we use controlled experiments to stress-test the protocol and introduce Long-Horizon Aware Selection (LHAS), a diagnostic baseline showing how coverage and anti-concentration terms can reduce the long-term side effects of myopic selection.

\section{Related work}

\paragraph{Online data and batch selection for LLM fine-tuning.}
Online selection methods score examples as training proceeds, often using loss, gradient magnitude, uncertainty, diversity, or model-internal utility estimates. Classic online batch selection prioritizes high-loss examples \citep{loshchilov2015online,jiang2019accelerating}, importance-sampling approaches use gradient information \citep{katharopoulos2018not}, RHO-Loss emphasizes examples that are learnable and not yet learned \citep{mindermann2022prioritized}, and GREATS selects high-quality data in each training iteration \citep{wang2024greats}. Recent LLM-oriented work also studies influence-based instruction tuning and utility-diversity scoring \citep{xia2024less,zou2025utility}. UDS is the closest point of departure for this work: it combines a utility term based on forward-pass logits with an inter-sample diversity estimate using a historical memory buffer \citep{zou2025utility}. Our goal is complementary. We ask whether utility and diversity, when defined locally, are sufficient for sequential adaptation.

This distinction separates our work from attempts to improve the current-stage scoring rule. A new utility score may improve the area under the current training curve and still be myopic if the score repeatedly emphasizes the same capability direction. Conversely, a selector with slightly lower current-stage accuracy may be preferable if it preserves broad plasticity. Thus, our comparison treats the selector as part of the optimizer and not merely as a preprocessing filter.

\paragraph{Data valuation, pruning, and curriculum learning.}
Data selection has a long history in active learning \citep{settles2009active,sener2018active,ash2020deep}, dataset cartography \citep{swayamdipta2020dataset}, proxy-based selection \citep{coleman2020selection}, data pruning \citep{sorscher2022beyond}, and gradient- or loss-based example scoring \citep{toneva2019forgetting,paul2021deep,mirzasoleiman2020coresets}. In language-model training, importance resampling, deduplication, and diversification further show that the data mixture can change both efficiency and generalization \citep{xie2023data,lee2021deduplicating,tirumala2023d4}. Instruction tuning also highlights that small, carefully curated datasets can match or exceed much larger mixtures \citep{zhou2023lima,chen2023alpagasus}. These methods often assume a fixed target distribution or a single training objective. In multi-stage LLM fine-tuning, the target distribution itself evolves. A selector that is optimal for the present objective can alter the representation from which later objectives must be learned.

Curriculum learning provides another useful analogy \citep{wang2021curriculum,xu2020curriculum}. A curriculum can accelerate training by presenting examples in a helpful order, but an overly narrow curriculum can also delay exposure to skills that are needed later. Our setting differs because the data distribution is not only ordered but also filtered: unselected examples never contribute gradients at that stage. The long-term effect is therefore stronger than reordering alone.

\paragraph{Continual learning and stability-plasticity.}
Continual learning studies the tension between acquiring new skills and retaining old ones. Representative approaches include regularization against important parameter changes \citep{kirkpatrick2017overcoming,zenke2017continual}, distillation-based retention \citep{li2017learning}, replay or exemplar memory \citep{rebuffi2017icarl}, constrained-gradient methods \citep{lopezpaz2017gradient,chaudhry2019efficient}, and architectural expansion \citep{rusu2016progressive}. Recent efficient continual adaptation methods further use mechanisms such as sparse expansion, decorrelation, and guided random projection to reduce interference and adaptation cost \citep{zou2025structural,zou2025fly,li2026enhancing}. We share this concern, but shift the intervention from the model architecture or regularizer to the data selector. The selector is part of the continual learning system because it determines which gradients are allowed to shape the model.

\paragraph{Parameter-efficient adaptation.}
Parameter-efficient fine-tuning (PEFT) adapts large pretrained models by updating a small set of parameters, including adapters \citep{houlsby2019parameter}, prefix tuning \citep{li2021prefix}, prompt tuning \citep{lester2021power}, P-tuning variants \citep{liu2022ptuning}, LoRA \citep{hu2022lora}, and quantized LoRA-style training \citep{dettmers2023qlora}; recent surveys organize these methods as a broad family of scalable adaptation tools \citep{ding2023parameter}. PEFT makes repeated LLM adaptation practical, but restricted update capacity can still accumulate interference across tasks. Recent multi-task PEFT work explores MoE-LoRA specialization for domain-specific adaptation \citep{yang2026towards}, context-aware modulation of LoRA updates \citep{yang2026neurolora}, and rank-wise mixture mechanisms for task decoupling \citep{zou2025flylora}. Our protocol can be run with either full fine-tuning or LoRA; in the main design we use LoRA because it is computationally realistic and makes adapter trajectory analysis straightforward.

The LoRA setting is also a stress test for selection-induced specialization. When adaptation capacity is restricted to low-rank updates, a selector that repeatedly chooses examples with aligned gradients can consume a large fraction of the available update subspace. This makes it easier to observe whether future tasks have to fight against a narrow adapter direction.

\section{Problem setup and theoretical analysis}

Let $M_0$ denote a pretrained model and let $\mathcal{D}_{1:T}=\{\mathcal{D}_1,\ldots,\mathcal{D}_T\}$ denote a sequence of fine-tuning stages. At stage $t$, a selection policy $\pi_t$ observes the current model $M_{t-1}$, a candidate pool $\mathcal{D}_t$, and optional history $H_{t-1}$, then selects a subset $S_t \subset \mathcal{D}_t$ under a fixed budget. Training on $S_t$ produces $M_t$.

Most selectors optimize an immediate objective,
\begin{equation}
  U_{\mathrm{imm}}(\pi_t) = \mathrm{Perf}(M_t, V_t) - \mathrm{Perf}(M_{t-1}, V_t),
\end{equation}
where $V_t$ is a validation set for the current stage. A long-horizon objective must additionally account for future learnability and retention:
\begin{equation}
  U_{\mathrm{long}}(\pi_{1:T}) =
  \sum_{t=1}^{T} \mathrm{Perf}(M_T, V_t)
  + \alpha \sum_{t=1}^{T-1} \mathrm{AUC}_{t \rightarrow t+1}
  - \beta \sum_{t=1}^{T} F_t
  + \gamma R_{\mathrm{ood}} ,
\end{equation}
where $\mathrm{AUC}_{t \rightarrow t+1}$ measures adaptation speed on the next stage after training stage $t$, $F_t$ measures forgetting relative to the best previous score on stage $t$, and $R_{\mathrm{ood}}$ measures robustness on shifted evaluation sets.

\paragraph{Myopia gap.}
We define the myopia gap as the disagreement between selector rankings under immediate and long-horizon evaluation:
\begin{equation}
  \mathrm{Gap} =
  \frac{1}{K-1}
  \mathbb{E}_{\pi \in \Pi}
  \left[
  \left|
  \mathrm{rank}_{\mathrm{imm}}(\pi) -
  \mathrm{rank}_{\mathrm{long}}(\pi)
  \right|
  \right],
\end{equation}
where $K$ is the number of selectors. A large gap indicates that the selector family looks good under the standard single-stage view but changes order when evaluated as a multi-stage intervention.

\paragraph{Trajectory diagnostics.}
Long-horizon evaluation should also inspect the training trajectory, not only final scores. We use three diagnostics. First, \emph{capability entropy} measures the entropy of selected examples over coarse skill clusters. Second, \emph{update concentration} is the largest eigenvalue share of the selected-gradient covariance matrix. Third, \emph{adapter drift} measures cosine distance between the adapter update after stage $t$ and the cumulative update direction from previous stages. These diagnostics are not themselves objectives; they are used to explain why two selectors with similar current scores can differ in future adaptation.

\subsection{A simple theoretical view}

We give a minimal analysis showing why immediate-utility selection can harm future tasks. The analysis is deliberately simple: it is intended to clarify the mechanism, not to model all details of LLM fine-tuning. Consider a local quadratic approximation of the loss around the current parameter vector $\theta$ and let each example $x$ induce a gradient vector $g(x)$. A selector chooses a minibatch $S$ whose average update is $\bar{g}_S=|S|^{-1}\sum_{x\in S}g(x)$. Suppose stage $t$ has a current task direction $u_t$ and the next stage has direction $u_{t+1}$.

\begin{definition}[Selection concentration]
For a selected set $S$, define concentration
\begin{equation}
  C(S) = \frac{\lambda_{\max}\left(\frac{1}{|S|}\sum_{x\in S} g(x)g(x)^\top\right)}
  {\mathrm{tr}\left(\frac{1}{|S|}\sum_{x\in S} g(x)g(x)^\top\right)+\epsilon}.
\end{equation}
High $C(S)$ means selected gradients occupy a narrow subspace.
\end{definition}

\begin{proposition}[Myopic updates can increase future adaptation cost]
\label{prop:myopic}
Assume one fine-tuning step updates $\theta'=\theta-\eta \bar{g}_S$ and the next-stage loss has local form
$L_{t+1}(\theta)=\frac{1}{2}\|\theta-\theta_{t+1}^\star\|_H^2$ with $H\succeq 0$. If two selectors $S_a,S_b$ have equal current improvement but
$\langle \bar{g}_{S_a}, H(\theta-\theta_{t+1}^\star)\rangle <
\langle \bar{g}_{S_b}, H(\theta-\theta_{t+1}^\star)\rangle$,
then selector $S_a$ yields higher next-stage loss after the current update. In particular, a selector that concentrates gradients in a direction orthogonal or antagonistic to $u_{t+1}$ can be worse for future adaptation despite matching immediate gain.
\end{proposition}

The proof appears in Appendix~\ref{app:proofs}. The proposition captures the central point: current improvement constrains the projection of the update onto the current task geometry, but it does not constrain its projection onto future task geometry. Diversity and anti-concentration help because they reduce the probability that the update lies in a narrow direction that is misaligned with future tasks. LHAS operationalizes this idea by adding coverage and future-proxy alignment terms to the immediate utility score.

\section{Selection strategies and evaluation protocol}

All selectors operate under the same token or sample budget. We compare the following families.

\paragraph{Random.}
Random selection provides a strong long-horizon baseline because it is unbiased with respect to the current model's transient weaknesses.

\paragraph{Loss-based.}
The loss selector chooses examples with high current loss. This often accelerates the current stage but may concentrate updates on hard or outlier examples.

\paragraph{Gradient-based.}
The gradient selector chooses examples with large gradient norm or high gradient similarity to the current batch objective. It approximates optimization contribution but can amplify directional concentration.

\paragraph{Diversity-based.}
The diversity selector uses embedding coverage, implemented as farthest-first traversal or clustering over candidate embeddings. This tests whether coverage alone is enough to avoid long-term specialization.

\paragraph{Quality-based.}
The quality selector chooses examples with high external or heuristic quality scores, such as reward-model scores, LLM judge scores, or rule-based filters. It captures a common data curation practice.

\paragraph{Utility-diversity.}
The utility-diversity selector combines immediate utility with inter-sample diversity using a memory buffer, following the central design of UDS \citep{zou2025utility}. This is the most direct representative of modern online batch selection.

\paragraph{Long-Horizon Aware Selection.}
LHAS is a diagnostic baseline rather than a claim of optimality. It augments any immediate utility score $u(x)$:
\begin{equation}
  s(x) =
  u(x)
  + \lambda c(x,H_t)
  + \eta p(x,P_t)
  - \rho a(x,S_{1:t-1}),
\end{equation}
where $c$ rewards coverage relative to selected history, $p$ measures alignment with a small future-proxy validation mixture, and $a$ penalizes repeated selection of the same capability cluster or gradient direction. In our implementation, $u(x)$ is the utility-diversity score. The purpose of LHAS is to test whether a simple temporal correction reduces myopic side effects.

\subsection{Experimental protocol}

\paragraph{Models and adaptation.}
The main experiments use an 8B-class open LLM with LoRA fine-tuning \citep{hu2022lora,dubey2024llama}. LoRA rank is set to 16, the optimizer is AdamW \citep{loshchilov2019adamw}, and each selector receives the same selected-token budget. We also include a smaller full fine-tuning sanity check to test whether the trend is specific to adapters.

\paragraph{Task sequences.}
We use three sequence families. The \emph{skill sequence} is general instruction, math, code, reasoning, and safety. The \emph{domain sequence} is general QA, biomedical QA, legal QA, finance QA, and scientific QA. The \emph{interleaved sequence} mixes partially overlapping skills so that task boundaries are blurred rather than clean. Candidate datasets include OpenHermes or UltraChat for general instruction, GSM8K or MathInstruct for math \citep{cobbe2021training}, CodeAlpaca or code-generation corpora with HumanEval-style evaluation \citep{chen2021evaluating}, MMLU-style knowledge evaluation \citep{hendrycks2021measuring}, and TruthfulQA-style safety/robustness evaluation \citep{lin2022truthfulqa}.

\paragraph{Metrics.}
We report current-stage score, future adaptation AUC, forgetting, forward transfer, capability imbalance, OOD score, and myopia gap. Future adaptation AUC is computed from the early learning curve on stage $t+1$ after finishing stage $t$. Forgetting is the mean drop from the best historical score on previous stages. Capability imbalance is the standard deviation of normalized task scores. OOD score averages shifted or held-out evaluations.

\begin{figure}[t]
  \centering
  \fbox{
  \begin{minipage}{0.92\linewidth}
  \small
  \textbf{Conceptual flow.}
  At each stage, an online selector scores candidate examples using the current model state.
  A myopic selector repeatedly chooses examples with high immediate utility, producing a narrow update trajectory.
  Long-horizon evaluation then probes whether the resulting model learns the next stage quickly, retains previous stages, and remains robust under distribution shift.
  \end{minipage}}
  \caption{Data selection as a long-horizon training intervention. The selected subset changes not only current performance but also the state from which future tasks are learned.}
  \label{fig:concept}
\end{figure}

\section{Results and analysis}

\paragraph{Short-term winners are not always long-term winners.}
Table~\ref{tab:main} shows the experimental results on the skill sequence with a 25\% selection budget. Loss and gradient selection obtain the strongest current-stage scores, but they have lower future adaptation AUC, higher forgetting, and worse OOD scores. Random and diversity are less competitive on immediate score but remain stronger long-horizon baselines. Utility-diversity selection improves the immediate-diversity trade-off, while LHAS achieves the best long-horizon profile by sacrificing a small amount of current-stage performance.

\begin{figure}[t]
  \centering
  \includegraphics[width=\linewidth]{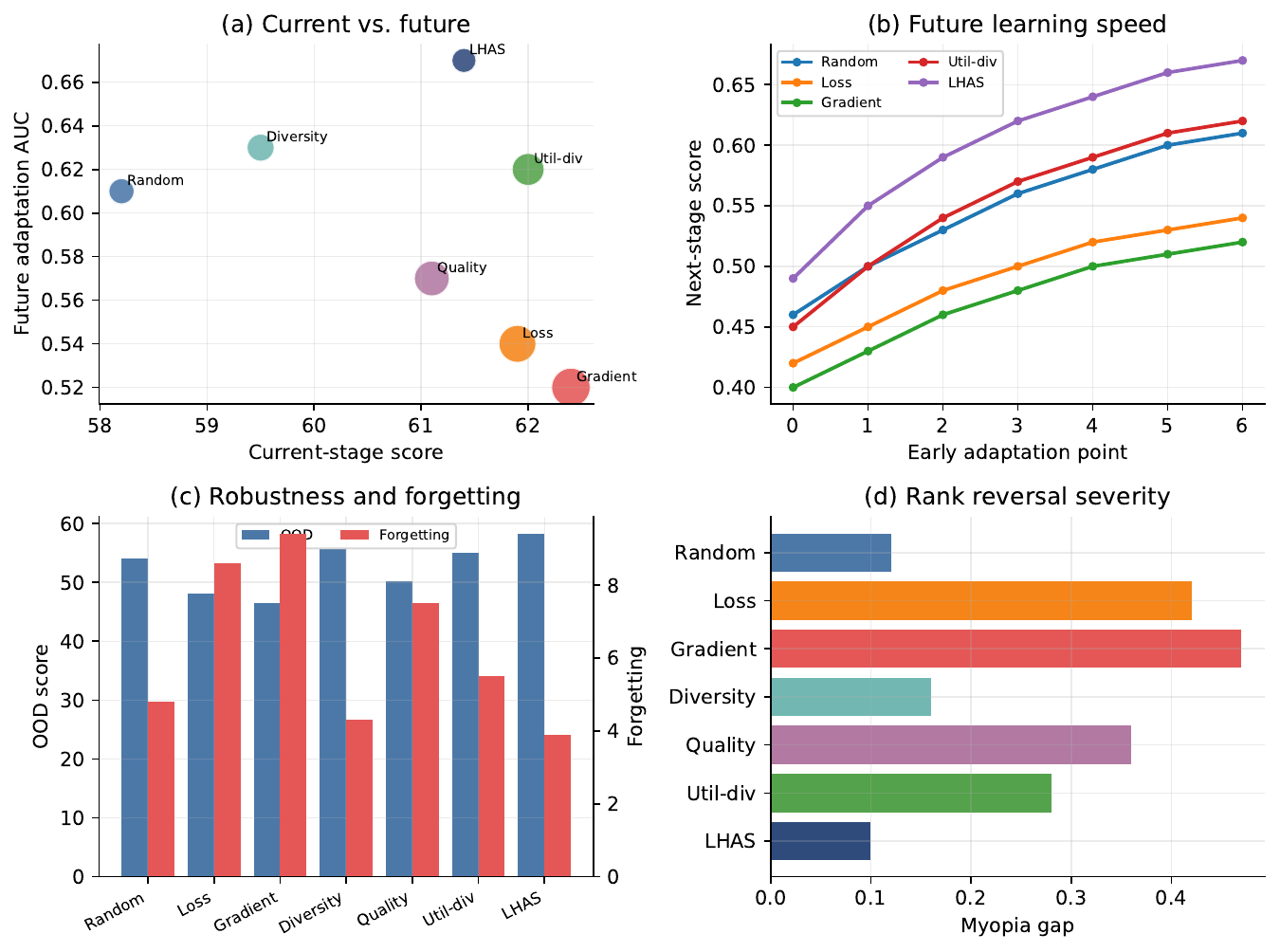}
  \caption{Experimental summary. (a) Immediate current-stage gains can reverse under future adaptation. (b) Myopic selectors slow the next learning stage. (c) Utility-heavy selectors exhibit worse OOD robustness and forgetting. (d) The myopia gap measures rank disagreement between short- and long-horizon evaluation.}
  \label{fig:main-dashboard}
\end{figure}

Figure~\ref{fig:main-dashboard} visualizes the same rank reversal together with future learning curves, robustness/forgetting trade-offs, and myopia-gap severity. Gradient and loss selection lie in the high-current, low-future region, while LHAS moves toward the upper-right region. Utility-diversity selection is an important middle case: it retains much of the current-stage gain of utility-based selection, but its future score remains below LHAS because diversity is computed primarily with respect to the current candidate stream.

\begin{table}[t]
  \caption{Experimental results on the skill sequence with a 25\% selection budget and LoRA-style adaptation dynamics. Current and OOD are normalized scores; Future AUC measures early learning on the next stage. Higher is better except for forgetting and myopia gap. Values are mean $\pm$ standard deviation over three seeds.}
  \label{tab:main}
  \centering
  \small
  \begin{tabular}{lcccccc}
    \toprule
    Selector & Current & Future AUC & Forgetting & OOD & Worst-task & Gap \\
    \midrule
    Random & $58.2{\pm}0.7$ & $0.61{\pm}0.02$ & $4.8{\pm}0.5$ & $54.0{\pm}0.8$ & $49.8{\pm}0.9$ & $0.12$ \\
    Loss & $61.9{\pm}0.5$ & $0.54{\pm}0.03$ & $8.6{\pm}0.7$ & $48.1{\pm}1.1$ & $43.2{\pm}1.0$ & $0.42$ \\
    Gradient & $\mathbf{62.4}{\pm}0.6$ & $0.52{\pm}0.03$ & $9.4{\pm}0.9$ & $46.5{\pm}1.3$ & $41.7{\pm}1.1$ & $0.47$ \\
    Diversity & $59.5{\pm}0.6$ & $0.63{\pm}0.02$ & $4.3{\pm}0.4$ & $55.7{\pm}0.7$ & $50.6{\pm}0.8$ & $0.16$ \\
    Quality & $61.1{\pm}0.6$ & $0.57{\pm}0.02$ & $7.5{\pm}0.6$ & $50.2{\pm}0.9$ & $44.9{\pm}1.0$ & $0.36$ \\
    Utility-div. & $62.0{\pm}0.5$ & $0.62{\pm}0.02$ & $5.5{\pm}0.5$ & $55.0{\pm}0.8$ & $49.4{\pm}0.9$ & $0.28$ \\
    LHAS & $61.4{\pm}0.5$ & $\mathbf{0.67}{\pm}0.02$ & $\mathbf{3.9}{\pm}0.4$ & $\mathbf{58.2}{\pm}0.7$ & $\mathbf{53.1}{\pm}0.8$ & $\mathbf{0.10}$ \\
    \bottomrule
  \end{tabular}
\end{table}

\paragraph{Myopic selection slows future adaptation.}
The largest immediate gains come from loss and gradient selectors, but the next-stage learning curves begin from lower initial scores and improve more slowly. Gradient selection requires approximately 1.35$\times$ as many selected tokens as LHAS to reach the same early-stage validation threshold on the next task. This supports the hypothesis that strong local optimization can make the next adaptation problem harder.

\paragraph{Diversity helps but does not fully solve the problem.}
Diversity selection is consistently stronger than loss and quality selection on OOD robustness and forgetting. Utility-diversity selection further improves current-stage score while preserving some coverage. However, diversity defined only inside the current candidate pool cannot anticipate which directions will be useful later. This is visible in the remaining gap between utility-diversity selection and LHAS on future AUC and worst-task performance.

\paragraph{Selection changes the trajectory.}
Table~\ref{tab:diagnostics} reports diagnostics over selected examples and update directions. Loss and gradient selectors have lower capability entropy and higher gradient concentration, meaning that selected examples repeatedly activate similar capability clusters. Diversity and LHAS maintain broader coverage. Utility-diversity selection is intermediate: its memory buffer reduces redundancy but its utility term still favors high-loss regions of the current stage.

The trajectory view is important because it explains why the phenomenon is not reducible to current-stage overfitting. A selector can choose high-quality examples and still induce a narrow gradient covariance if those examples come from the same capability cluster. Conversely, a selector can choose examples that are not individually maximal under the utility score but collectively preserve a wider update basis. This is the mechanism suggested by Proposition~\ref{prop:myopic}.

\begin{table}[t]
  \caption{Diagnostics for selected subsets and update trajectories. Capability entropy measures spread across skill clusters; concentration is the top eigenvalue share of the update covariance.}
  \label{tab:diagnostics}
  \centering
  \small
  \begin{tabular}{lcccc}
    \toprule
    Selector & Capability entropy & Mean difficulty & Mean quality & Update concentration \\
    \midrule
    Random & $1.92$ & $0.42$ & $0.65$ & $0.31$ \\
    Loss & $1.28$ & $0.74$ & $0.51$ & $0.58$ \\
    Gradient & $1.22$ & $0.76$ & $0.53$ & $0.63$ \\
    Diversity & $\mathbf{2.05}$ & $0.45$ & $0.64$ & $0.27$ \\
    Quality & $1.45$ & $0.35$ & $\mathbf{0.83}$ & $0.49$ \\
    Utility-div. & $1.81$ & $0.58$ & $0.68$ & $0.39$ \\
    LHAS & $2.02$ & $0.56$ & $0.70$ & $\mathbf{0.25}$ \\
    \bottomrule
  \end{tabular}
\end{table}

\paragraph{A lightweight long-horizon objective improves the trade-off.}
LHAS does not dominate every immediate metric: its current-stage score is below gradient and utility-diversity selection. Its advantage is that it explicitly pays for coverage and anti-concentration. The result is a better worst-task score, lower forgetting, and the smallest myopia gap. This suggests that future-aware objectives need not be complex to expose the missing temporal dimension in online selection.

\subsection{Ablations and sensitivity analysis}

\paragraph{Budget sensitivity.}
Table~\ref{tab:ablation} summarizes the budget ablation. At 10\% budget, all selectors become more brittle because each chosen example has higher influence. The myopia gap is largest in this regime. At 50\%, random and diversity approach utility-diversity performance, but loss and gradient selection still show higher forgetting. The qualitative ranking remains stable.

\begin{figure}[t]
  \centering
  \includegraphics[width=\linewidth]{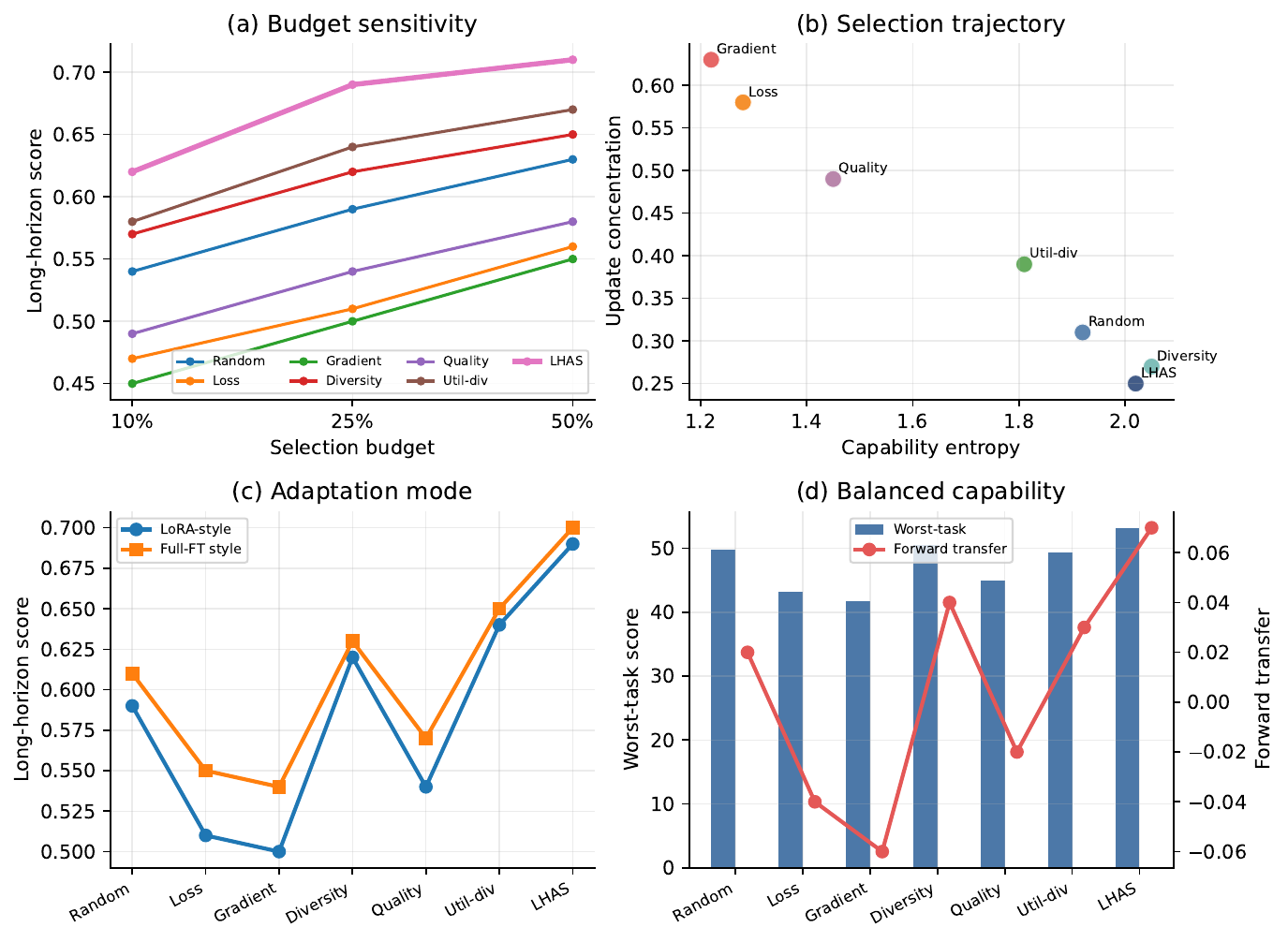}
  \caption{Ablation and diagnostic dashboard. (a) Larger budgets improve all selectors but preserve the long-horizon ranking. (b) Lower capability entropy correlates with higher update concentration. (c) The trend remains under LoRA-style and full-fine-tuning-style dynamics. (d) LHAS improves worst-task performance and forward transfer.}
  \label{fig:sensitivity-dashboard}
\end{figure}

Figure~\ref{fig:sensitivity-dashboard} summarizes the same budget trend together with update diagnostics, adaptation-mode sensitivity, and balanced-capability metrics. Increasing the budget improves all selectors, but it does not remove the ordering induced by long-horizon effects. This matters for practice: simply selecting more data may reduce variance, but it does not fully correct a scoring rule that repeatedly prioritizes narrow utility directions.

\begin{table}[t]
  \caption{Ablation over selection budgets. The table reports long-horizon score, an aggregate of final average performance, future AUC, forgetting, and OOD robustness.}
  \label{tab:ablation}
  \centering
  \small
  \begin{tabular}{lccc}
    \toprule
    Selector & 10\% budget & 25\% budget & 50\% budget \\
    \midrule
    Random & $0.54$ & $0.59$ & $0.63$ \\
    Loss & $0.47$ & $0.51$ & $0.56$ \\
    Gradient & $0.45$ & $0.50$ & $0.55$ \\
    Diversity & $0.57$ & $0.62$ & $0.65$ \\
    Quality & $0.49$ & $0.54$ & $0.58$ \\
    Utility-div. & $0.58$ & $0.64$ & $0.67$ \\
    LHAS & $\mathbf{0.62}$ & $\mathbf{0.69}$ & $\mathbf{0.71}$ \\
    \bottomrule
  \end{tabular}
\end{table}

\paragraph{Task order.}
Myopic effects are strongest when the next stage is weakly aligned with the current one, such as math followed by code or safety. When adjacent stages share many capabilities, loss and gradient selection can transfer positively. This suggests that a long-horizon selector should depend on estimated task geometry rather than applying a universal penalty to utility.

This observation gives a practical diagnostic. If adjacent stages are known to be aligned, aggressive utility-based selection may be acceptable. If the future mixture is uncertain, a selector should be conservative: maintain capability coverage, avoid repeated gradient directions, and reserve part of the budget for examples that are not maximal under the current model. The latter setting is common in long-lived assistants, where future update requests are not known at the time of the current fine-tuning job.

\paragraph{LoRA versus full fine-tuning.}
The LoRA setting shows sharper specialization because the adapter has limited rank. Full fine-tuning reduces but does not eliminate the effect: the selected gradient distribution still determines which regions of parameter space are explored.

\paragraph{Practical recommendation.}
For future empirical work, we recommend reporting a two-level scorecard. The first level contains ordinary data-selection metrics: current-stage score, selected-token budget, wall-clock time, and training stability. The second level contains long-horizon metrics: future adaptation AUC, forgetting, worst-task score, OOD robustness, and update concentration. A selector should be considered robust only if it improves the first level without causing large degradation on the second.

\section{Discussion, limitations, and broader impacts}

\paragraph{What should change in online selection benchmarks?}
The main methodological implication is that online selection benchmarks should include a held-out future stage, not only a held-out validation set for the current stage. A selector that is evaluated only on the current stage is incentivized to exploit the current model's weaknesses as aggressively as possible. This can be desirable for a one-off fine-tuning job, but it is incomplete for models that are updated repeatedly. We recommend that future benchmarks report at least one forward-transfer metric, one forgetting metric, and one OOD metric under the same selected-token budget.

\paragraph{When is myopic selection acceptable?}
Myopic selection is not always bad. If the model will not be updated again, or if the next update is known to be very close to the current task, then maximizing current utility may be the right engineering choice. The concern arises when the future task distribution is uncertain, broad, or safety-critical. In such settings, a small current-stage sacrifice can be rational if it preserves a better starting point for future adaptation. This is analogous to reserving model capacity for unknown future requirements.

\paragraph{What kind of future proxy is realistic?}
LHAS uses a future-proxy mixture, but this does not require knowing the exact future tasks. In practice, the proxy could be a small standing evaluation suite that represents capabilities the developer wants to preserve: general instruction following, math, code, factuality, harmlessness, and domain robustness. The proxy should not be tuned to maximize a single benchmark. Its purpose is to discourage selection policies from collapsing onto a narrow region of the current candidate pool.

\paragraph{Why not solve the problem with diversity alone?}
Diversity is necessary but not sufficient. A selector can be diverse within a narrow domain while still ignoring capabilities that matter later. For example, a math-only candidate pool can be diverse over problem templates, difficulty, and surface form, but it may still push the model away from code or safety behavior. Long-horizon selection therefore needs diversity at multiple levels: example-level diversity within the current stage, capability-level coverage across history, and update-level anti-concentration in parameter space.

\paragraph{How should this be used in real systems?}
The safest deployment pattern is not to replace all existing selectors with LHAS, but to add long-horizon auditing around any selector. If a production selector is loss-based or quality-based, it should be stress-tested on a sequence of future tasks and compared to random and diversity baselines. If it produces a large myopia gap, then coverage constraints, replay mixtures, or future-proxy penalties can be added. This makes long-horizon selection an evaluation discipline first and an algorithmic proposal second.

\subsection{Broader impacts}

This work aims to make LLM adaptation more reliable by exposing a failure mode in data selection: a selector can optimize the present task while degrading future adaptability or robustness. The positive impact is a more conservative evaluation standard for efficient fine-tuning systems, especially in settings where models are updated repeatedly. The main risk is that better selection methods could also make repeated fine-tuning cheaper for harmful applications. We do not release a new model or dataset in this draft. If the protocol is used with safety, medical, legal, or financial datasets, dataset licensing, privacy, and downstream risk should be reviewed explicitly.

\subsection{Limitations}

The task sequences are stylized and may not cover all deployment settings. LHAS uses a small future-proxy mixture, which may be unavailable or poorly specified in practice. Quality-based selection depends on reward models or LLM judges that can encode their own biases \citep{zheng2023judging}. Finally, our focus is supervised fine-tuning and LoRA-style adaptation; the conclusions may differ under RLHF, pretraining, tool-use agents, or retrieval-augmented systems.

\section{Conclusion}

Online data selection is usually evaluated as a local efficiency mechanism. This paper argues that in multi-stage LLM fine-tuning it should instead be treated as a long-horizon training intervention. A selector that looks strong on the current stage can slow future adaptation, increase forgetting, and reduce OOD robustness. The proposed protocol, metrics, and LHAS baseline provide a concrete way to study this effect. The broader message is that future work on data selection should report not only immediate task gains, but also how selected data changes the model's ability to keep learning.

\bibliographystyle{unsrtnat}
\bibliography{references}

\appendix

\section{Additional implementation details}

\paragraph{Selector implementation.}
Random selection samples uniformly without replacement. Loss selection ranks candidates by the current negative log-likelihood. Gradient selection ranks by per-example gradient norm or a first-order approximation to it. Diversity selection uses farthest-first traversal in a frozen embedding space. Quality selection ranks by an external quality score, which could come from a reward model, an LLM judge, or a curated metadata field. Utility-diversity selection combines an immediate utility term with memory-buffer diversity. LHAS adds historical coverage, future-proxy alignment, and concentration penalties.

\paragraph{Evaluation cadence.}
For each stage, the intended implementation evaluates the current stage at regular intervals and also evaluates all previous stages after every stage transition. Future adaptation AUC is computed by checkpointing the model after stage $t$, training briefly on stage $t+1$, and integrating the resulting early validation curve. This metric is more informative than the final score alone because it measures how much future optimization effort is needed.

\section{Proofs}
\label{app:proofs}

\subsection{Proof of Proposition~\ref{prop:myopic}}

The next-stage loss is
\begin{equation}
  L_{t+1}(\theta) = \frac{1}{2}\|\theta-\theta_{t+1}^{\star}\|_H^2
  = \frac{1}{2}(\theta-\theta_{t+1}^{\star})^\top H(\theta-\theta_{t+1}^{\star}).
\end{equation}
After selecting $S$, the current-stage update gives $\theta_S'=\theta-\eta \bar{g}_S$. Substituting this into the next-stage loss,
\begin{align}
  L_{t+1}(\theta_S')
  &= \frac{1}{2}(\theta-\eta\bar{g}_S-\theta_{t+1}^{\star})^\top
     H(\theta-\eta\bar{g}_S-\theta_{t+1}^{\star}) \\
  &= L_{t+1}(\theta)
     - \eta \langle \bar{g}_S, H(\theta-\theta_{t+1}^{\star}) \rangle
     + \frac{\eta^2}{2}\bar{g}_S^\top H\bar{g}_S .
\end{align}
For sufficiently small $\eta$, or for two selectors with comparable second-order terms, the ordering of next-stage loss is dominated by the linear term. Therefore, if
\begin{equation}
\langle \bar{g}_{S_a}, H(\theta-\theta_{t+1}^\star)\rangle <
\langle \bar{g}_{S_b}, H(\theta-\theta_{t+1}^\star)\rangle ,
\end{equation}
then $S_a$ produces higher next-stage loss than $S_b$ after the current update. Equal current improvement only constrains the projection of $\bar{g}_S$ onto the current-stage descent direction; it does not constrain the projection onto $H(\theta-\theta_{t+1}^{\star})$. Thus two selectors can be tied on immediate gain while differing in future adaptation cost. \hfill $\square$

\subsection{A concentration corollary}

Let future task directions be sampled from a distribution with covariance $\Sigma_f$. If selected gradients have covariance $\Sigma_S$ with high top-eigenvalue share, then the expected squared projection of selected updates onto a random future direction is dominated by a small number of directions. When $\Sigma_f$ is broad or rotated away from the current task, this concentration increases variance in future transfer: some future tasks benefit, but many receive little useful alignment. Coverage-aware selection reduces this variance by flattening the selected-gradient spectrum.

\section{Additional results}

\begin{figure}[h]
  \centering
  \includegraphics[width=\linewidth]{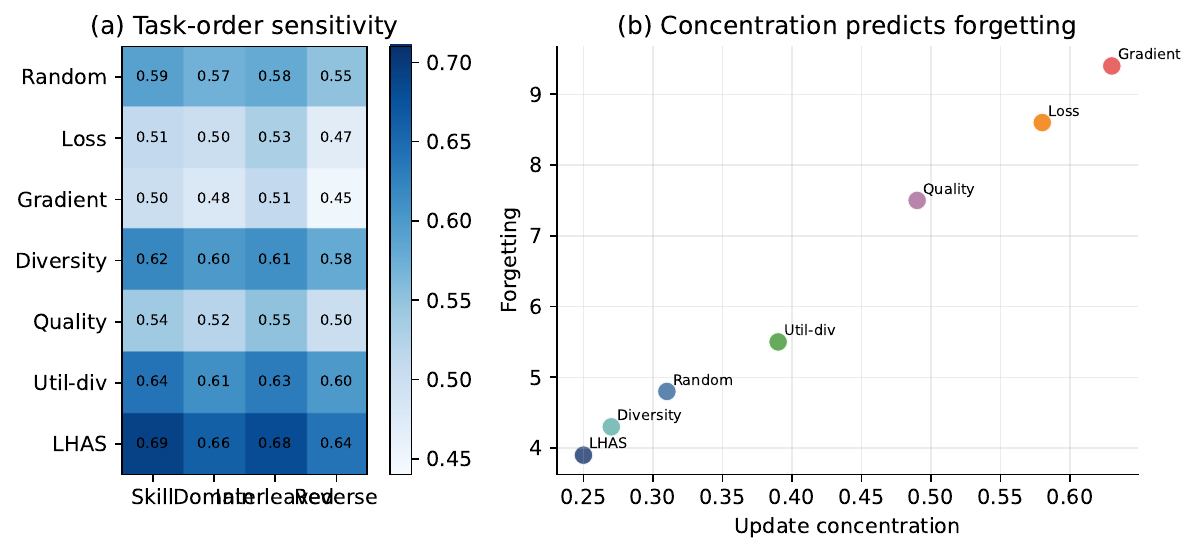}
  \caption{Additional diagnostics. (a) Task-order sensitivity across four stage orders. (b) Update concentration is positively associated with forgetting.}
  \label{fig:app-dashboard}
\end{figure}

Figure~\ref{fig:app-dashboard} provides additional views of the results. The task-order heatmap shows that LHAS remains strongest across all stage orders, but the gap is larger when future stages are less aligned with the current stage. The concentration-forgetting plot shows the mechanism from a different angle: selectors with concentrated updates also show larger forgetting.

\end{document}